\documentclass[acmsmall,nonacm=true,screen=true]{acmart}

\AtBeginDocument{%
  \providecommand\BibTeX{{%
    \normalfont B\kern-0.5em{\scshape i\kern-0.25em b}\kern-0.8em\TeX}}}

\setcopyright{acmcopyright}
\copyrightyear{202x}
\acmYear{202x}
\acmDOI{10.1145/123123123.121212}

\acmConference[arXiv]{arXiv '2x}{date}{arXiv}
\acmBooktitle{arXiv '2x, date, arXiv}
\acmPrice{00.00}
\acmISBN{978-1-4503-XXXX-X/2x/xx}

\usepackage{subfig}
\usepackage{tikz}
\usetikzlibrary{positioning, calc, shapes.geometric, shapes.symbols, arrows.meta, fit, backgrounds, intersections}
\usepackage{pgfplots}
\pgfplotsset{compat=1.16}
\usepackage{pgfplotstable}

\usepackage{xstring}

\begin{document}

\title{AI-Supported Assessment of Load Safety}

\author{Julius Schöning}
\email{j.schoening@hs-osnabrueck.de}
\affiliation{%
  \institution{Osnabrück University of Applied Sciences}
  \department{Faculty of Engineering and Computer Science}
  \city{Osnabrück}
  \country{Germany}
  \postcode{DE-49076}
}

\author{Niklas Kruse}
\email{niklas.kruse@hs-osnabrueck.de}

\affiliation{%
  \institution{Osnabrück University of Applied Sciences}
  \department{Faculty of Engineering and Computer Science}
  \city{Osnabrück}
  \country{Germany}
  \postcode{DE-49076}
}

\renewcommand{\shortauthors}{}
\renewcommand{\shorttitle}{}

\begin{abstract}
Load safety assessment and compliance is an essential step in the corporate process of every logistics service provider. In 2020, a total of 11,371 police checks of trucks were carried out, during which 9.6\% (1091) violations against the load safety regulations were detected. For a logistic service provider, every load safety violation results in height fines and damage to reputation. An assessment of load safety supported by artificial intelligence (AI) will reduce the risk of accidents by unsecured loads and fines during safety assessments. This work shows how photos of the load, taken by the truck driver or the loadmaster after the loading process, can be used to assess load safety. By a trained two-stage artificial neural network (ANN), these photos are classified into three different classes I) cargo loaded safely, II) cargo loaded unsafely, and III) unusable image. By applying several architectures of convolutional neural networks (CNN), it can be shown that it is possible to distinguish between unusable and usable images for cargo safety assessment. This distinction is quite crucial since the truck driver and the loadmaster sometimes provide photos without the essential image features like the case structure of the truck and the whole cargo. A human operator or another ANN will then assess the load safety within the second stage.
\end{abstract}








\maketitle

\section{Introduction}
The assessment of load safety is an important step in the business process of any logistic service provider. In 2020, 11,371 traffic checks were carried out, during which 1091 violations against load safety were detected \cite{BLM2020}. Load safety violations are accountable for almost one-tenth of all violations and often result in high fines \cite{BundesamtJustiz2023}. Both the driver and the loadmaster are liable in the event of a violation under current case law. In addition to penalties imposed by authorities, civil claims may also arise if outside parties are adversely affected \cite{Sander2023}. In this respect, it is of enormous importance for the logistics service provider to inspect the load safety, reducing their liability risk properly.

In order to prove that they are in line with the legal compliances of load safety, logistic companies often take photos of the cargo in the trucks as an internal safety check before the trucks leave the warehouses. Some logistic companies inspect the photos by highly qualified quality management employees and allow the truck's departure only after successful control. However, due to a large number of loads, this kind of follow-up inspection by a quality management employee is very time-consuming and may not always be possible due to the low image quality of the photos due to the circumstances of image acquisition in the warehouse by the driver or the loadmaster.

This work investigates to what extent artificial intelligence (AI) can be used to improve or facilitate the work of the quality management employee in assessing load safety based on photos. Therefore the photos should be classified into three different classes I) cargo loaded safely, II) cargo loaded unsafely, and III) unusable image. By applying several architectures of convolutional neural networks (CNN), it can be shown that it is possible to distinguish between unusable and usable images. This distinction is quite crucial since the truck driver and the loadmaster sometimes provide photos without the essential image features like the case structure of the truck and the whole cargo. The load safety will then be assessed by a human operator or another artificial neural network (ANN). With such an AI-based safety assessment system in place, the legal certainty of a logistics company will increase, and the bottleneck, the quality management department, will vanish.

\section{State of the Art}
Starting with the big picture, this section introduces a centralized system for load safety assessment, then gives an overview of ANN for image classification. Going even deeper, this section will also describe standard techniques to improve and prepare the datasets for ANN by data augmentation.

\subsection{Centralized System for Load Safety Assessment}
Digitalizing logistics, processes, and data of trucks, cargo, traffic flow etc., will be managed in a centralized platform. This platform could be a cloud run by a cloud service provider or a cloud run on-premise, i.e., on an in-house IT solution. Having a centralized platform available, this platform could be used to verify cargo safety.

Systems for cargo safety verification are essential since not all of a company's drivers are permanent employees, yet the loadmaster is legally liable for violations by these drivers \cite{Sander2023}. The cargo is first loaded by a subcontractor and then inspected by a permanent company employee. First, any dangerous goods are checked to see if they were loaded safely, and then the rest of the load is examined. If both checks are successful, the employee takes a photo of the load and sends it via a centralized platform to quality management. Using a centralized platform that not only collects photos of the load but also uses AI to examine the safety of the load could significantly reduce the workload in both quality management and the warehouse.

\subsection{Artificial Neural Networks for Image Classification}
ANN can be used for a wide variety of tasks like classifying objects in images \cite{Abiodun2018}, recognizing speech and words \cite{Schmidhuber2015}, creating CAD-models from images \cite{Schoening2015},  and controlling mechatronic systems \cite{Schoening2022}, to only name a few. In general, ANN can recognize different features of a dataset \cite{Choi2020,Schmidhuber2015}. Features on taken images could be, e.g., the edges of a piece of cargo, the framework of the truck, the floor of the truck, and the colors of labels. During the supervised training of an ANN, the ANN selected the used features on its own. There are different architectures of ANN. For image classification tasks like the load safety assessment, CNN performs well. CNN are more efficient compared to other architectures for recognizing and learning the relevant features of the image itself \cite{OShea2015}.

CNN in the field of image classification have evolved rapidly in the past, so that many models exist today, using the convolution technique \cite{Khan2020}. Since these architectures are freely available, it seems reasonable to identify architectures as state-of-the-art and then use them for load safety assessment. CNN architectures such as VGG\cite{Simonyan2014}, AlexNet \cite{Krizhevsky2017}, GoogleNet \cite{Szegedy2017}, and ResNet \cite{He2016} are commonly used. However, in many application cases, the size of the input image for the CNN architecture is often not optimally chosen. Allowing an optimally designed CNN architecture, Richter et al. \cite{Richter2021} provide a Python toolbox for analyzing the architecture of CNN by receptive fields. An overview table of the minimum and maximum feasible input image resolution on Tensorflow and Torch built-in CNN implementation is given in \cite{Richter2022,Schoening2023}

In the training of ANN, the dataset is of particular importance since it was shown in the literature \cite{AbdElrahman2013,Kaur2019} that better results can be achieved if the data of the individual classes are balanced. If this is not the case, there is the problem that individual classes are under or overrepresented, and the overfitting effect occurs.

\subsection{Data Augmentation}
In object classification, data augmentation describes the artificial extension of a dataset by using image processing techniques \cite{Khalifa2021,Taylor2018}. Data augmentation should be used when an overfitting effect appears within the training of the ANN \cite{Khalifa2021}. Overfitting usually occurs because there is not enough data in the dataset, so the ANN does not have the opportunity to learn in a generalized manner. The collection of new data can be very costly \cite{Taylor2018} because, on the one hand, the data must first be reassigned to specific classes, and, on the other hand, new data can only be found with considerable effort in a short time.

The group of classical data augmentation techniques is divided into subgroups of geometric and photometric processing \cite{Khalifa2021,Taylor2018}. Geometric data augmentation changes only the spatial properties of the image and only, to a limited extent, the actual representation. The most straightforward and commonly used image processing technique is to make a copy of the image using a flip on the x or y-axis. The application of geometric image processing techniques must be performed in a very targeted manner \cite{Khalifa2021}. By randomly cropping an original image, relevant features may be lost, negatively affecting learning. Thus, to achieve the desired effect, it must be carefully investigated beforehand whether the processed images still have a proper relation to the original domain \cite{Khalifa2021,Shorten2019}.

\section{Image-Based Load Safety Classification}
For assessing load safety, the ANN must be able to classify the images into three different classes I) cargo loaded safely, II) cargo loaded unsafely, and III) unusable image.
This classification task could be solved in two ways. First, a multi-class classification directly classifies whether the images show either safe cargo, unsafe cargo, or an unusable image. An alternative way is to divide the classification task into a decision tree of two binary classifications. Thus the first branch of the decision tree assesses whether the input image is usable or unusable. In the case of a usable image, a distinction between cargo loaded safely or unsafely is made, respectively.

By visual inspection of the used dataset, the class III) unusable image is easily distinguishable from the classes I) cargo loaded safely and II) unsafely, even for a non-trained human observer. It can therefore be assumed that using a decision tree with two binary classifications might also be easier for an ANN. The recognition of unsafely and safely loaded cargo seems more complex at first sight. However, it is currently carried out by humans in the quality management department of a logistics company; thus, an ANN might be able to perform this task too. Nevertheless, solving only the first classification step to identify unusable images will already have considerable value for any logistic service provider.

\subsection{Structure of the Dataset}
The used dataset is composed of 5712 images. These images were taken during the shipping process of a logistic company and always showed the truck's rear view. These images were taken by employees of the logistic company who check the safety of the load while still at the loading center and ultimately create and transmit a photo for quality assurance.
This real-world dataset is taken since, in the context of AI, the transferability of the academic datasets to real-world industrial deployment is often quite challenging \cite{Wuest2016,Lee2020}. Using a dataset as close as possible to the target domain can simplify the usability of the results \cite{Chen2021}. The advantage of the dataset is that it does not come from an outside source but was used for evaluation at a logistic company and manually classified into the relevant classes. At the same time, however, this classification could lead to errors in the initial classification, which, given the dataset size, could be attributed to human error or different standards in the assessment \cite{Northcutt2021}. Nevertheless, this should facilitate the subsequent transfer of AI findings to real-life applications in the field.

With some exceptions, the images have a resolution of 3456x4608 pixels and RGB color space. In advance, the images were divided into the three classes mentioned above.

The I) cargo loaded safely class is formed by all images that show suitable load safety and represent the optimum result from the logistics company's point of view. There are 1813 images in this class.

The II) cargo loaded unsafely class contains the images that were previously described as safe in the warehouse and before the departure of the load but would be considered unsafe in a subsequent inspection. Thus, images in this class do not satisfy legal requirements and could carry significant liability risks \cite{BundesamtJustiz2023,Sander2023}. Examples of this class represent the least desirable case, as they could give rise to liability risks. In this respect, this class must be reliably detected to allow a logistics company to defend itself against fines. This class consists of 2355 images.

The III) unusable image class is characterized by the fact that they are not suitable for checking cargo security by quality management. In other words, no visible features in the images can provide information about the safety of the goods because they are too blurred, too dark, or do not show the entire cargo hold, for example. There are 1544 images in this class.

\subsection{Applied Data Augmentation}
Data augmentation aims to prevent possible overfitting and thus improve the training effect, whereby one reason for overfitting is a too-small dataset \cite{Khalifa2021,Shorten2019}. However, the number of datasets collected within the logistics company is limited and cannot be easily expanded with original data due to the significant effort involved in the new survey. Initial pre-run tests found that overfitting occurs very early in the training process with the given data.

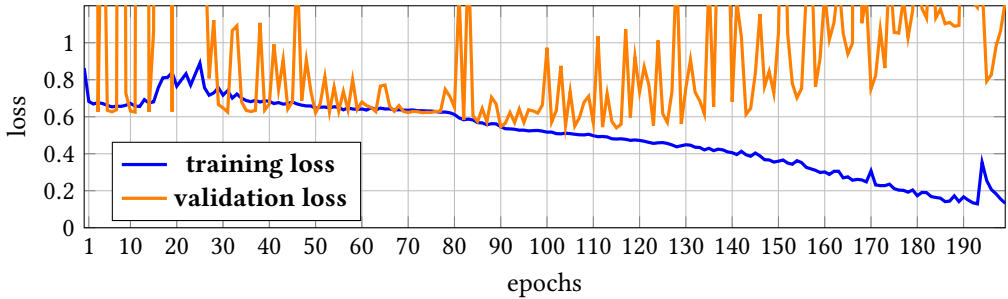
\begin{figure}[!t]
\centering

  \pgfplotstableread[col sep = comma]{plots/InceptionV3200.csv}\datatable

  {
  \begin{tikzpicture}
  \begin{axis}[
      width=0.995\linewidth,
      height=4.5cm,
      ymin=0,
      ymax=1.2,
      ytick={0,0.2,...,1},
      xmin=0,
      xmax=199,
      xtick={1,10,20,...,200},
      xmajorgrids,
      ymajorgrids,
      legend pos=south west,
      xlabel={epochs},
      ylabel={loss},
  ]
  \addplot [blue, very thick] table [x ={epoch}, y={loss}]{\datatable};
    \addlegendentry{\textbf{training loss}}
  \addplot [orange, very thick] table [x ={epoch}, y={valloss}]{\datatable};
      \addlegendentry{\textbf{validation loss}}
  \end{axis}

\end{tikzpicture}}

\caption{Training and validation loss gradient of the InceptionV3 architecture, overfitting occurs approximated at epoch 50.}
\label{fig:loss}
\end{figure}

The preliminary test with the InceptionV3 architecture showed, as illustrated in Fig. \ref{fig:loss} that the overfitting effect occurred approximated at epoch 50. From this epoch, the model lost its generalized competence and emphasized the need for data augmentation. The remaining networks showed comparable effects in the same preliminary tests before they reached the desired metrics in terms of accuracy.

For the implementation of data augmentation, photometric and geometric methods are used. By not altering the classes of the images, only the random flip on the y axes, slight random rotation, random brightness modification, and random color adjustments are used for data augmentation.

\subsection{Architecture Design of the Convolutional Neural Network}
Three CNN architectures are used for the experiment, which can be assigned to two categories.  On the one hand, two deep ANN were used for the InceptionV3 \cite{Szegedy2015} and the ResNet101 \cite{He2016}. On the other hand, one shallow architecture, called LogisticNet based on the AlexNet architecture \cite{Krizhevsky2017}, is used.

In the standardized design, InceptionV3 and the ResNet101 have an input resolution of 299x299 \cite{Szegedy2015,Szegedy2017}, and 244x244 \cite{He2016} pixels, respectively. The resolution can significantly impact the model's performance, so a higher resolution can also lead to better classification accuracy \cite{Kannojia2018}. The resolution to be chosen can be further influenced by the receptive field of an ANN \cite{Richter2021,Richter2022}. Since the receptive field can be calculated, for some state-of-the-art models, there are defined maximum and minimum values that the input image should have to utilize the existing model optimally. For the ResNet101 this resolution is a maximum of 971x971 pixels; for the InceptionV3, it is 1311x1311 pixels. Due to hardware limitations and to allow batch sizes larger than two, the resolution here was capped at 800x800 pixels.

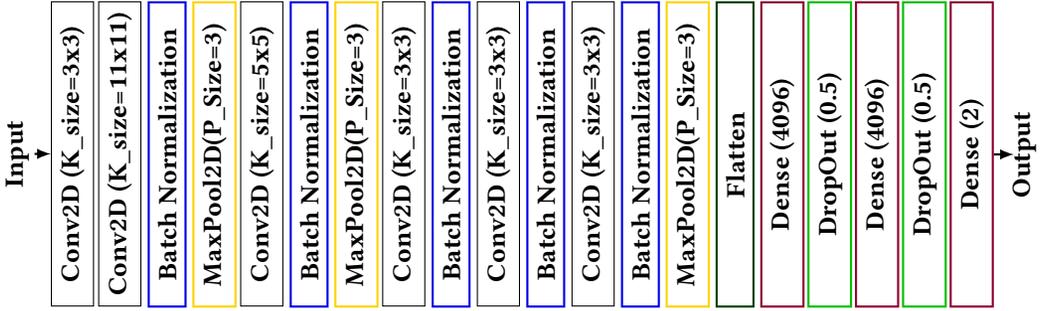
\begin{figure}[!t]
\centering
\begin{tikzpicture}
  \node(input)[draw=none, rotate=90, rectangle, minimum width = 4.0cm, minimum height = 0.3cm] {\textbf{Input}};
  \node(c1)[right = 0.73cm of input, anchor = east, draw, rotate=90, rectangle, minimum width = 4.0cm, minimum height = 0.3cm] {\textbf{Conv2D (K\_size=3x3)}};
  \node(c2)[right = 0.63cm of c1, anchor = east, draw, rotate=90, rectangle, minimum width = 4.0cm, minimum height = 0.3cm] {\textbf{Conv2D (K\_size=11x11)}};
  \node(B1)[right = 0.63cm of c2, anchor = east, draw=blue!90!black, rotate=90, thick, rectangle, minimum width = 4.0cm, minimum height = 0.3cm] {\textbf{Batch Normalization}};
  \node(MP)[right = 0.63cm of B1, anchor = east, draw=yellow!85!red, rotate=90, thick, rectangle, minimum width = 4.0cm, minimum height = 0.3cm] {\textbf{MaxPool2D(P\_Size=3)}};
  \node(c3)[right = 0.63cm of MP, anchor = east, draw, rotate=90, rectangle, minimum width = 4.0cm, minimum height = 0.3cm] {\textbf{Conv2D (K\_size=5x5)}};
  \node(B2)[right = 0.63cm of c3, anchor = east, draw=blue!90!black, rotate=90, thick, rectangle, minimum width = 4.0cm, minimum height = 0.3cm] {\textbf{Batch Normalization}};
  \node(MP2)[right = 0.63cm of B2, anchor = east, draw=yellow!85!red, rotate=90, thick, rectangle, minimum width = 4.0cm, minimum height = 0.3cm] {\textbf{MaxPool2D(P\_Size=3)}};
  \node(c4)[right = 0.63cm of MP2, anchor = east, draw, rotate=90, rectangle, minimum width = 4.0cm, minimum height = 0.3cm] {\textbf{Conv2D (K\_size=3x3)}};
  \node(B3)[right = 0.63cm of c4, anchor = east, draw=blue!90!black, rotate=90, thick, rectangle, minimum width = 4.0cm, minimum height = 0.3cm] {\textbf{Batch Normalization}};
  \node(c4)[right = 0.63cm of B3, anchor = east, draw, rotate=90, rectangle, minimum width = 4.0cm, minimum height = 0.3cm] {\textbf{Conv2D (K\_size=3x3)}};
  \node(B3)[right = 0.63cm of c4, anchor = east, draw=blue!90!black, rotate=90, thick, rectangle, minimum width = 4.0cm, minimum height = 0.3cm] {\textbf{Batch Normalization}};
  \node(c4)[right = 0.63cm of B3, anchor = east, draw, rotate=90, rectangle, minimum width = 4.0cm, minimum height = 0.3cm] {\textbf{Conv2D (K\_size=3x3)}};
  \node(B3)[right = 0.63cm of c4, anchor = east, draw=blue!90!black, rotate=90, thick, rectangle, minimum width = 4.0cm, minimum height = 0.3cm] {\textbf{Batch Normalization}};
  \node(MP3)[right = 0.63cm of B3, anchor = east, draw=yellow!85!red, rotate=90, thick, rectangle, minimum width = 4.0cm, minimum height = 0.3cm] {\textbf{MaxPool2D(P\_Size=3)}};
  \node(F1)[right = 0.63cm of MP3, anchor = east, draw=green!25!black, rotate=90, thick, rectangle, minimum width = 4.0cm, minimum height = 0.3cm] {\textbf{Flatten}};
  \node(D1)[right = 0.63cm of F1, anchor = east, draw=purple!75!black, rotate=90, thick, rectangle, minimum width = 4.0cm, minimum height = 0.3cm] {\textbf{Dense (4096)}};
  \node(Do1)[right = 0.63cm of D1, anchor = east, draw=green!75!black, rotate=90, thick, rectangle, minimum width = 4.0cm, minimum height = 0.3cm] {\textbf{DropOut (0.5)}};
  \node(D1)[right = 0.63cm of Do1, anchor = east, draw=purple!75!black, rotate=90, thick, rectangle, minimum width = 4.0cm, minimum height = 0.3cm] {\textbf{Dense (4096)}};
  \node(Do1)[right = 0.63cm of D1, anchor = east, draw=green!75!black, rotate=90, thick, rectangle, minimum width = 4.0cm, minimum height = 0.3cm] {\textbf{DropOut (0.5)}};
  \node(D1)[right = 0.63cm of Do1, anchor = east, draw=purple!75!black, rotate=90, thick, rectangle, minimum width = 4.0cm, minimum height = 0.3cm] {\textbf{Dense (2)}};
  \node(Out)[right = 0.7cm of D1, anchor = east, draw=none, rotate=90, thick, rectangle, minimum width = 4.0cm, minimum height = 0.3cm] {\textbf{Output}};
  \draw [-latex, thick ](D1) -- ($(Out.north)+(0.15,0)$);
  \draw [-latex, thick ]($(input.south)-(0.05,0)$) -- (c1);
  \end{tikzpicture}
\caption{LogisticNet architecture based on the AlexNet \cite{Krizhevsky2017} architecture.}
\label{fig:logiNet}
\end{figure}

The LogisticNet, shown in Fig. \ref{fig:logiNet} is based on the AlexNet, which itself is purely sequential \cite{Simonyan2014}. For this reason, the input resolution was limited to 227x227 pixels \cite{Krizhevsky2017}. LogisticNet intends to find out whether a simple model is also capable of solving the task, so a comprehensive redesign of LogisticNet was omitted.

However, the LogisticNet has some adaptations compared to the AlexNet. In its first convolution layer, the AlexNet has a large kernel size of 11x11 \cite{Krizhevsky2017}. Newer models tend to use smaller kernels here \cite{Khan2020,Novakovic2017}, as they act more efficiently and are better at extracting small features from the dataset \cite{Camgoezlue2021}. The remaining convolution layers are oriented based on the original AlexNet architecture, but the last dense layer reduces the number of classes to two classes instead of 1000. As with the other models, the softmax activation function was chosen as the activation function of the last dense layer.

\section{Evaluation}
All three architectures are trained throughout 300 epochs. Since it is expected that they would show the effects of overfitting at a certain point, an epoch had to be found at which the training was terminated before overfitting occurred, and the model can no longer recognize general features.

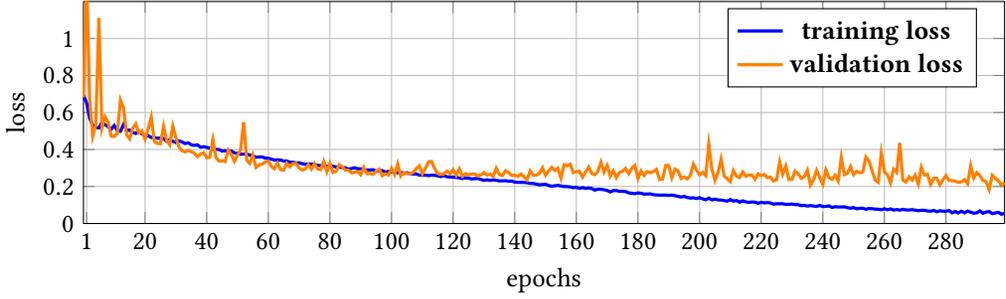
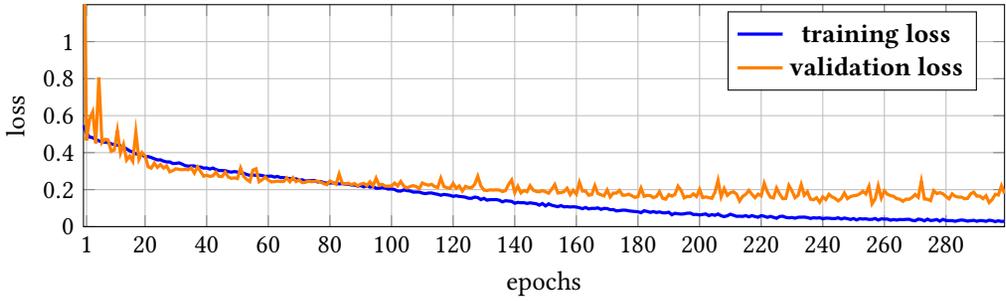
\begin{figure}[!t]
\centering
\subfloat[ResNet101 \label{sub:ResNet18}] {
\centering
\pgfplotstableread[col sep = comma]{plots/ResNet800x800Q300V12.csv}\datatable
\begin{tikzpicture}
\begin{axis}[
    width=0.995\linewidth,
    height=4.5cm,
    ymin=0,
    ymax=1.2,
    ytick={0,0.2,...,1},
    xmin=0,
    xmax=299,
    xtick={1,20,40,...,300},
    xmajorgrids,
    ymajorgrids,
    legend pos=north east,
    xlabel={epochs},
    ylabel={loss},
]
\addplot [blue, very thick] table [x ={epoch}, y={loss}]{\datatable};
  \addlegendentry{\textbf{training loss}}
\addplot [orange, very thick] table [x ={epoch}, y={valloss}]{\datatable};
    \addlegendentry{\textbf{validation loss}}
\end{axis}
\end{tikzpicture}}

\subfloat[InceptionV3 \label{sub:Inception}] {
\centering
\pgfplotstableread[col sep = comma]{plots/InceptionV3800x800Q300V12.csv}\datatable
\begin{tikzpicture}
\begin{axis}[
    width=0.995\linewidth,
    height=4.5cm,
    ymin=0,
    ymax=1.2,
    ytick={0,0.2,...,1},
    xmin=0,
    xmax=299,
    xtick={1,20,40,...,300},
    xmajorgrids,
    ymajorgrids,
    legend pos=north east,
    xlabel={epochs},
    ylabel={loss},
]
\addplot [blue, very thick] table [x ={epoch}, y={loss}]{\datatable};
  \addlegendentry{\textbf{training loss}}
\addplot [orange, very thick] table [x ={epoch}, y={valloss}]{\datatable};
    \addlegendentry{\textbf{validation loss}}
\end{axis}
\end{tikzpicture}}

\caption{Loss curves of \protect\subref{sub:ResNet18} ResNet101 and \protect\subref{sub:Inception} InceptionV3 for the input resolution 800x800px.}
\label{fig:lossResnet}
\end{figure}

The ResNet101 achieved a validation accuracy of 87\% at the resolution of 244x244 after 300 epochs and reached its highest validation accuracy at epoch 242 with a validation accuracy of 93\%. On the resolution of 800x800px, the validation accuracy was 94\% after 300 epochs and reached its highest point after 292 epochs with 94.5\%. As plotted in Fig. \ref{fig:lossResnet} the loss curves of both models, the overfitting effect is noticeable. At the 244x244px resolution, the ResNet101 reached the point of overfitting after 90 epochs, while at the 800x800px resolution, the overfitting did not occur until epoch 120. For the final evaluation, ResNet101 was trained at a resolution of 244x244px over 87 epochs and on 800x800px over 117 epochs avoiding the overfitting on the training dataset.

The InceptionV3 achieved a validation accuracy of 95\% after 300 epochs in the native resolution of 299x299px and a value of 97\% in the high resolution of 800x800px. It reached its high point with an accuracy of 96\% at epoch 277 in the low resolution and 96\% after 280 epochs in the high resolution. The previously observed overfitting effect is also seen in InceptionV3. Here, the overfitting occurred in the test with the low resolution after epoch 58. In the higher resolution test, the overfitting could be observed from epoch 85. In the following test, the InceptionV3 network was trained in low resolution over 55 epochs and high resolution over 82 epochs.

The LogisticNet was only tested in a resolution of 227x227px, so only one version of the model must be evaluated. After a test duration of 300 epochs, the LogisticNet achieved a validation accuracy of 88\%. The LogisticNet reached the highest point of validation accuracy after 254 epochs with a value of 89\%. The overfitting effect was visible in the LogisticNet from epoch 49. The LogisticNet was trained over 46 epochs in a second run and then evaluated.

\begin{table}[!t]
\centering
\caption{Metrics of the distinction between usable, class I) as well as II), and unusable images class III) by the different ANN architectures.}
\begin{tabular}{|cccc|}
\hline
\multicolumn{4}{|c|}{{ ResNet101}}                                                                                           \\ \hline
\multicolumn{2}{|c|}{Resolution: 244x244px}                               & \multicolumn{2}{c|}{Resolution: 800x800px}          \\ \hline\hline
\multicolumn{1}{|c|}{Recall: 94\%} & \multicolumn{1}{c|}{Precision: 91\%} & \multicolumn{1}{c|}{Recall: 92\%} & Precision: 89\% \\ \hline
\multicolumn{4}{|c|}{{ InceptionV3}}                                                                                         \\ \hline
\multicolumn{2}{|c|}{Resolution: 299x299px}                               & \multicolumn{2}{c|}{Resolution: 800x800px}          \\ \hline
\multicolumn{1}{|c|}{Recall: 94\%} & \multicolumn{1}{c|}{Precision: 92\%} & \multicolumn{1}{c|}{Recall: 93\%} & Precision: 90\% \\ \hline\hline
\multicolumn{4}{|c|}{{ LogisticNet}}                                                                                         \\ \hline
\multicolumn{2}{|c|}{Recall: \textbf{98\%}}                               & \multicolumn{2}{c|}{Precision: \textbf{95\%}}       \\ \hline
\end{tabular}
\label{tab:resutl}
\end{table}

\section{Result}
After finding the optimal count of training epochs for all networks, they were trained again, and the average results are shown in Tab. \ref{tab:resutl}. On distinguishing between usable and unusable  images, all CNN architecture achieves precision and recall above 90\%.

By applying, i.e., training and evaluating, the ResNet101 and the InceptionV3 architecture on the second task of the decision tree to distinguish between classes I) cargo loaded safely and II) cargo loaded unsafely, the archived precision and recall drastically decrease, as summarized in Tab. \ref{tab:resut2}.

\begin{figure}[!t]
\centering

%

\subfloat[299x299px \label{sub:low}] {
\includegraphics[width=.48\linewidth,trim={5cm 6.7cm 6cm 4.5cm},clip]{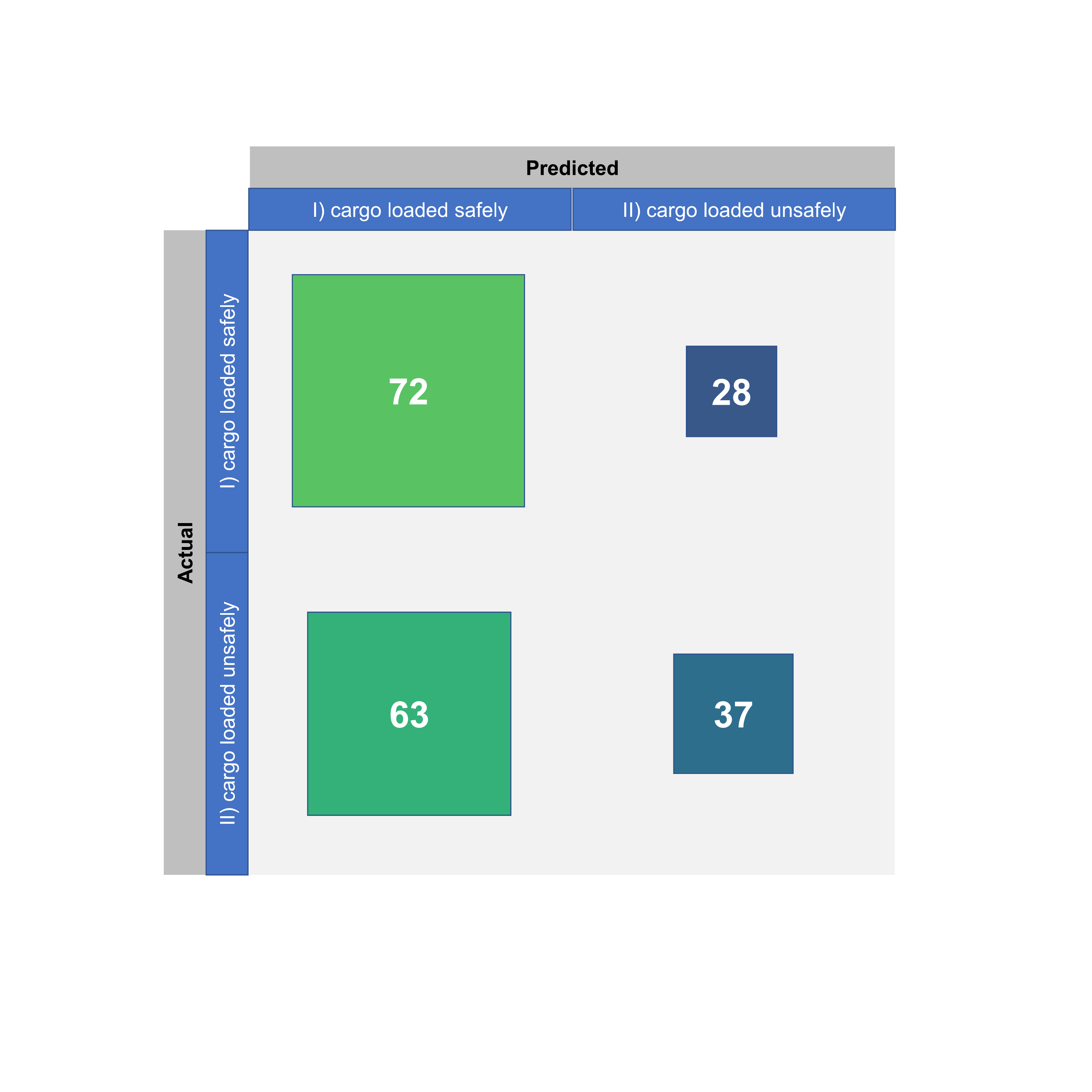}
}
\subfloat[800x800px \label{sub:high}] {
\includegraphics[width=.48\linewidth,trim={5cm 6.7cm 6cm 4.5cm},clip]{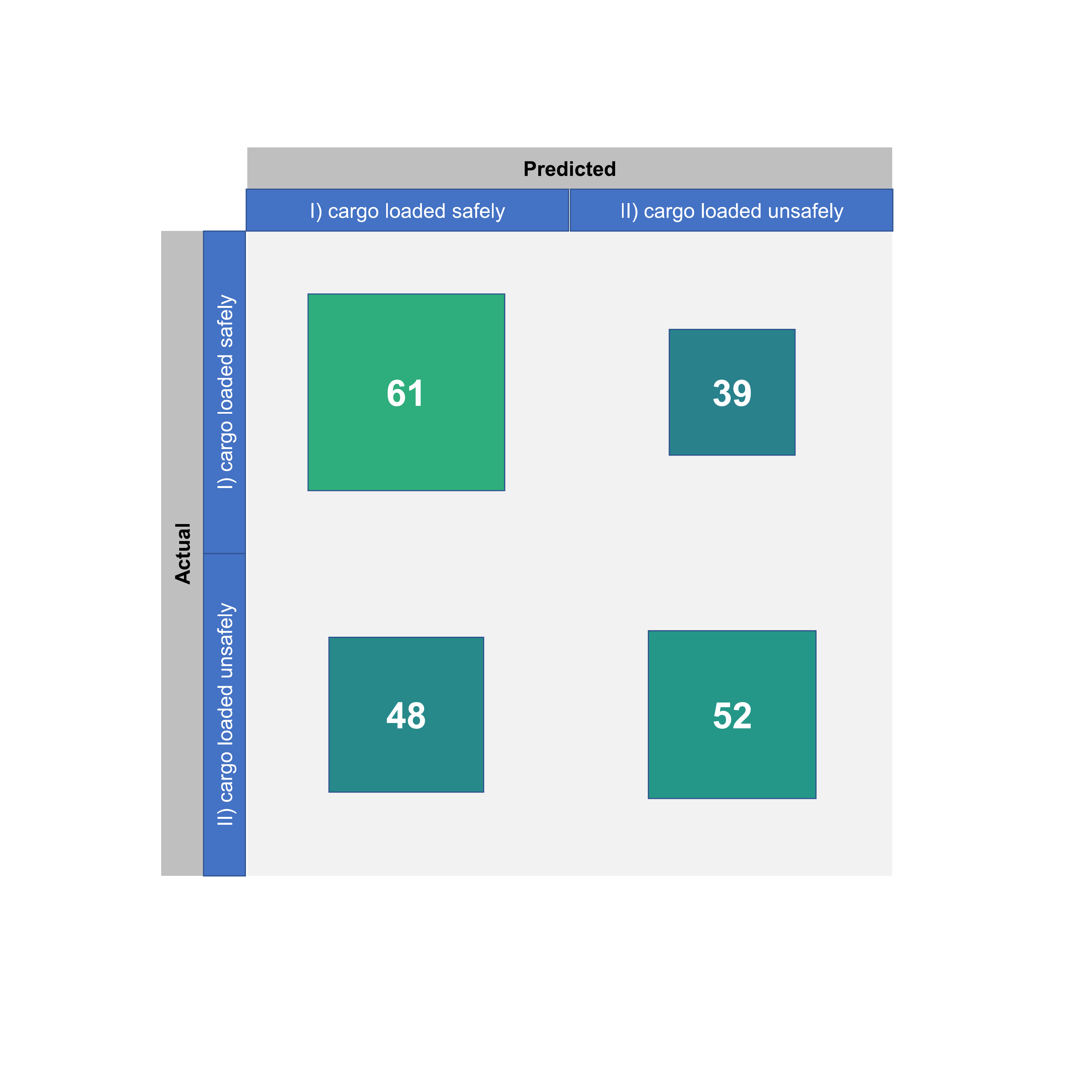}
}
\caption{Confusion Matrix distinguishing between classes I) cargo loaded safely and II) cargo loaded unsafely by Inception V3; resolution \protect\subref{sub:low} 299x299px and \protect\subref{sub:high} 800x800px.}
\label{fig:ICV3Q}
\end{figure}

As shown in the confusion matrix in Fig. \ref{fig:ICV3Q}, the InceptionV3 network, as the best architecture, was able to recognize 72 out of 100 images of class I) cargo loaded safely but only 37 out of 100 of the II) class, correctly. In the high resolution of 800x800px, different results were seen. Here 61 out of 100 images of class I) were recognized, and 51 out of 100 of class II) were correctly classified.

The ResNet101 could not achieve the same good results in this task's high and low resolution. In class I) only 45 out of 100 images were correctly classified, and in class II) 62 of 100 images were correctly recognized. Almost identical values were achieved in the high resolution, but the value of correctly classified results in class II) increased by three percent.

\begin{table}[!t]
\centering
\caption{Metrics of the distinction between classes I) cargo loaded safely and II) cargo loaded unsafely by ResNet101 and InceptionV3; MCC is the Matthews correlation coefficient where a 1.00 would represent a perfect classification.}
\begin{tabular}{|cccc|}
\hline
\multicolumn{4}{|c|}{ResNet101}                                                                                                           \\ \hline
\multicolumn{2}{|c|}{Resolution: 244x244px}                           & \multicolumn{2}{c|}{Resolution: 800x800px}                        \\ \hline
\multicolumn{1}{|c|}{F1-Score: 0.57} & \multicolumn{1}{c|}{MCC: 0.18} & \multicolumn{1}{c|}{F1-Score: \textbf{0.59}} & MCC: \textbf{0.20} \\ \hline\hline
\multicolumn{4}{|c|}{InceptionV3}                                                                                                         \\ \hline
\multicolumn{2}{|c|}{Resolution: 299x299px}                           & \multicolumn{2}{c|}{Resolution: 800x800px}                        \\ \hline
\multicolumn{1}{|c|}{F1-Score: 0.43} & \multicolumn{1}{c|}{MCC: 0.09} & \multicolumn{1}{c|}{F1-Score: 0.52}          & MCC: 0.13          \\ \hline
\end{tabular}

\label{tab:resut2}
\end{table}

\section{Conclusion}
Based on the result of the previous sections, AI-Supported assessment of load safety is feasible. It is valuable for any logistic service provider since a CNN can distinguish between usable and unusable images for cargo safety assessments. By implementing this function in the centralized logistic platform, the photos from the drivers and the loadmaster can automatically check if these photos are taken correctly for cargo safety assessment. This check will reduce the number of unusable images already during the acquisition of the images in the warehouse. Nevertheless, recognizing safe and unsafe loaded cargo could not be solved satisfactorily in this work. As seen in Tab. \ref{tab:resut2} and Fig. \ref{fig:ICV3Q}, the CNN architectures used could not sufficiently predict the images into the respective classes. One, but not the only, reason for this result could be the used dataset. By manual inspection of the results, an intersection between the classes I) cargo loaded safely and II) cargo loaded unsafely was notable. These intersections occur due to some boundary cases, where one cloud assesses the cargo shown on the image as both safe or unsafe loaded cargo. Overcoming this issue, the intersection of the classes must be removed. By removing these intersections, the used CNN architectures might generalize better between classes I) and II). In parallel, a hybrid dataset, a mixture of computer-generated and real-world images, will increase the number of training images and provide a basis for further work. In conclusion, AI can already support the assessment of load safety quite well but still need to be supervised by a human operator.
\bibliographystyle{myIEEEtran}
\bibliography{bib}

\end{document}